\documentclass{article}
\usepackage{spconf,amsmath,graphicx}
\usepackage{amssymb,amsmath,epsfig}
\usepackage{lipsum,color,multirow,url}
\usepackage{arydshln, algorithm, algpseudocode}
\usepackage{microtype}
\usepackage{nth}
\usepackage[inline]{enumitem}

\title{Low-Resource Contextual Topic Identification on Speech}
\makeatletter
\def\name#1{\gdef\@name{#1\\}}
\makeatother
\name{ \em Chunxi Liu$^{\dagger}$, Matthew Wiesner$^{\dagger}$, Shinji Watanabe$^{\dagger}$, Craig Harman$^{\dagger}$, Jan Trmal$^{\dagger,\ddagger}$, \\ \em Najim Dehak$^{\dagger}$, Sanjeev Khudanpur$^{\dagger,\ddagger}$  \thanks{This work was supported by DARPA LORELEI Grant N\b{o} HR0011-15-2-0024. The authors thank Zhongqiang Huang at the Raytheon BBN Technologies, USA, for his help with the machine translation systems.} }
\address{$^\dagger$Center for Language and Speech Processing, The Johns Hopkins University, USA \\
$^\ddagger$Human Language Technology Center of Excellence, The Johns Hopkins University, USA}
\begin{document}
\ninept
\maketitle
\begin{abstract}
In topic identification (topic ID) on real-world unstructured audio, an audio instance of variable topic shifts is first broken into sequential segments, and each segment is independently classified.
We first present a general purpose method for topic ID on spoken segments in low-resource languages, using a cascade of universal acoustic modeling, translation lexicons to English, and English-language topic classification.
Next, instead of classifying each segment independently, we demonstrate that exploring the contextual dependencies across sequential segments can provide large improvements. In particular, we propose an attention-based contextual model which is able to leverage the contexts in a selective manner.
We test both our contextual and non-contextual models on four LORELEI languages, and on all but one our attention-based contextual model significantly outperforms the context-independent models.
\end{abstract}
\begin{keywords}
Topic identification, universal acoustic modeling, recurrent neural networks, attention
\end{keywords}
%
\section{Introduction}
\label{sec:intro}

Storing and digitizing vast amounts of audio data such as broadcast news, telephone conversations, meetings, and lectures is now commonplace. However, to search, organize and analyze these large audio collections requires the development of new computational tools.
Topic identification (topic ID) from speech is one such human language technology that aims to identify topics or themes present in a speech recording.

Since audio data lacks the paragraphs and punctuation markings that naturally define semantically coherent chunks of text, long audio recordings of varying topic shifts must first be segmented, often manually.
Then the standard approach to topic ID from speech is to
\begin{enumerate*}[label=(\roman*)]
	\item develop automatic speech recognition (ASR) systems to decode each speech segment into word sequences,
   \item produce intermediate vector representations of the hypothesized word sequences for each segment, and
   \item learn a classifier from text/topic pairs and apply it to the vector representation of each segment independently.
\end{enumerate*}
However, such standard approach has many drawbacks, especially in a low-resource scenario: building ASR and topic classifiers in a new language requires a large amount of transcribed speech and topic-labeled text in the language, neither of which may be present. Furthermore, accurate topic inference or language understanding in general may require interpretation from adjacent segments. For instance, tasks such as anaphora resolution or entity disambiguation critically depend on contextual clues.

To study these challenges, we evaluate our topic ID performance in the DARPA LORELEI (Low Resource Languages for Emergent Incidents) Program framework. The program's goal is to develop human language technologies to support humanitarian assistance and disaster relief operations in locations where a low-resource language is spoken, also referred to as an incident language (IL) in the LORELEI terminology \cite{strassel2016lorelei,malandrakis2017extracting}. To provide situational awareness based on IL sources, one component task in LORELEI, called the Situation Frame (SF) task, involves building systems to provide meta-data for text and speech documents. These documents and associated meta-data are collectively referred to as situation frames (SFs) and consist of the following items:
\vspace{-0.1cm}
\begin{itemize}
\item situation type, also simply referred to as topic, 
\vspace{-0.1cm}
\item geographic localization, 
\vspace{-0.1cm}
\item status (temporal, resolution or urgency) of the situation. 
\end{itemize} 
\vspace{-0.1cm}
An SF system should automatically identify all the SFs covered in the text or speech collection of IL. In this paper we focus on building topic ID technology to enable situation type identification from speech.
\newcommand*{\myalign}[2]{\multicolumn{1}{#1}{#2}}
\begin{table*}[t]
\caption{\label{tab:sf_eg} {\it An example of a single spoken document that consists of seven spoken segments in LORELEI US English corpus.}}
\vspace{1mm}
\centering 
\begin{tabular}{  c     c    l   c   }
\hline \hline
 Document ID     &      Segment ID          &    \myalign{c}{Sampled sentences from spoken segment transcript}             &       Topic label         \\
\hline \hline
 USE\_080         &     USE\_080\_001       &    \multirow{2}{10.14cm} {turning to Tennessee where eleven people have now died in historic wildfires  ... hundreds of buildings have been torched   ... }    & Shelter    \\
                                      &                                                  &                &            \\
USE\_080         &      USE\_080\_002      &    \multirow{1}{10.14cm}{yeah you have a number of people missing but we don't know the exact number  ...  }     &     Out-of-domain      \\    
USE\_080          &      USE\_080\_003      &    \multirow{2}{10.14cm}{... and he said that the search and rescue effort yesterday ended and now today  it is search and recovery ... }   &  Urgent Rescue  \\  
                                      &                                                  &                &            \\
USE\_080          &      USE\_080\_004      &      ... so many homes damaged destroyed ...        &      Shelter       \\  
USE\_080          &      USE\_080\_005       &    \multirow{2}{10.14cm}{... just looking at the devastation now . because we saw a few homes and you know a few cars,  it is really bad ... }             &      Shelter       \\    
                                      &                                                  &                &            \\
USE\_080          &      USE\_080\_006      &     \multirow{2}{10.14cm}{... but people in town it sounds like now are questioning how fast they were notified to get out .... }        &    Evacuation        \\  
                                      &                                                  &                &            \\
USE\_080          &      USE\_080\_007      &     \multirow{2}{10.14cm}{... since they were forced to evacuate  so a lot of them will be seeing their homes and properties for the first time tonight ...}         &     Evacuation,        \\  
                                      &                                                  &                &   Shelter         \\
\hline \hline
\end{tabular}
\vspace{-0.6cm}
\end{table*}

\begin{table}[t]
\vspace{-0.2cm}
\caption{\label{tab:sf_topic} {\it Topic labels defined in the LORELEI Speech SF task.}}
\vspace{1mm}
\centering 
\begin{tabular}{ c  l  }
\hline \hline
    Topic scope  &      Topic label (Situation Type)                 \\
\hline \hline
                                            &      \multirow{1}{5.2cm}{Evacuation}     \\
                                          &          Food Supply                                                          \\    
                                           &          Urgent Rescue                                                    \\  
                                           &           Utilities, Energy, or Sanitation                                                      \\  
                                          &             Infrastructure                                                \\ 
           In-domain               &             Medical Assistance                                                    \\ 
                                          &            Shelter                                                  \\ 
                                          &          Water Supply                                                  \\ 
                                          &           Civil Unrest or Wide-spread Crime                                                             \\ 
                                          &              Elections and Politics                                               \\ 
                                          &            Terrorism or other Extreme Violence                                              \\ 
 \hline
  Out-of-domain                    &       Out-of-domain           \\
\hline \hline
\end{tabular}
\vspace{-0.7cm}
\end{table}

In order to simulate realistic disaster scenarios, the LORELEI speech corpora are divided into IL corpora -- corpora which typically contain unlabeled data in a low-resource language pertaining to one or more emergent disasters -- and related language corpora for which annotated data, possibly from high-resource languages, is provided. In both cases the audio data is collected ``in the wild'', and for a diverse set of languages. These data are collected, manually segmented, and annotated by APPEN \cite{appen} for the LORELEI program.
We refer to each unsegmented audio file as one spoken document. Since audio file segmentations are provided, each document consists of a sequence of segments, and each segment lasts around one minute on average and no more than 2 minutes.
There are 11 predefined topics chosen according to the LORELEI program scope, as shown in Table \ref{tab:sf_topic}. Any speech segment categorized by at least one of these topics is defined as in-domain data, otherwise as out-of-domain that can be viewed as the \nth{12} topic label. 
Table \ref{tab:sf_eg} shows an example spoken document that is split into 7 segments with varying topic. 

In this paper, we focus particularly on the IL scenario for which the only annotated data are from related (development) languages in addition to a very small amount of IL topic labeled data or IL transcribed speech (minutes rather than hours) which may be obtained. 

\section{Related Work}
\label{sec:relatedwork}

Prior work of topic ID on speech \cite{hazen2007topic, dredze2010nlp, wintrode2014limited, may2015topic} has focused on conversational telephone speech such as LDC's Fisher and Switchboard collections, where topic ID was performed for each whole conversation. Since the two participants of each conversation were prompted to speak on one single topic, no conversation segmentation was needed. Furthermore, since each conversation contains a single topic and lasts 5-10 minutes, the classification task is relatively straightforward.
The LORELEI collections, however, provide a more challenging and realistic scenario, where wildly collected audio recordings can be extremely long, of varying length, and contain multiple topic shifts at variable locations in the audio. For this reason each audio document in the LORELEI data is first segmented by APPEN \cite{appen}, and then topic ID is required on the much shorter resulting segments. 

To solve the LORELEI task, prior work \cite{papadopoulos2017team} used a mismatched ASR to directly decode IL speech, while \cite{wiesner2018jhu} proposed sharing common phonemic representation among languages and transferring acoustic models trained on higher-resource (potentially related) language(s). After ASR, \cite{papadopoulos2017team, wiesner2018jhu} translated both development (dev) and incident languages into English words, used the translated dev language data along with the given topic label annotations to learn English-language topic models and then classify the translated IL data.  
Additionally, instead of using ASR to convert speech into sequences of words, \cite{liu2017empirical, liu2017topic, wiesner2018jhu} also investigated unsupervised techniques to automatically discover and decode IL speech segments into phone-like units via acoustic unit discovery (AUD), or into word-like units via unsupervised term discovery (UTD). However,
only small amount of IL topic labels might be available to learn classifiers based on AUD/UTD tokenized segments, though \cite{wiesner2018jhu} showed marginal gains by combining them with the above cascade approach that implemented ASR, machine translation (MT) and operated on English words.

Extensive work on text classification has also been explored to date. 
For example, each word can be mapped to word embedding vector and concatenated as inputs to convolutional neural network \cite{kim2014convolutional} or recursive/recurrent neural networks \cite{socher2013recursive, tang2015document},  with a final softmax classification layer. 
\cite{yang2016hierarchical} demonstrated improved classification performance by using hierarchical attention networks to learn both word- and sentence-level attentions,  
enabling the models to attend differentially to more and less important words/sentences.   


However, all the above work is focused on classifying each data instance (i.e. each single sentence, conversation, or document) individually, and independently from the rest of data instances.  
Data instances in close proximity to each other may incorporate contextual information that can be exploited by contextual modeling. 
Thus we note that our LORELEI topic ID task, which is formulated as multi-label classification for each speech segment in a spoken document, is similar to the domain or intent classification in a multi-turn spoken language understanding (SLU) component of a dialog system \cite{xu2014contextual, liu2015deep, hori2015context}. One conversation session between user and dialog system, which can be viewed as one spoken document, may include multiple turns, and the user query in each turn is a spoken segment; thus, each segment needs to be classified into one of the supported domains or user intents, as classified into topic(s).
\cite{xu2014contextual, liu2015deep, hori2015context} have shown that SLU may require contextual interpretation from the dialog history, and the SLU models that incorporate the semantic contexts of preceding user utterances and system outputs outperform those without context. Therefore, in this paper, we investigate if the propagation of contextual information across spoken segments can improve topic ID, although the spoken segment that is one minute long on average in our case is often much longer and more semantically self-contained than the typical utterance of a few words in SLU systems.

\section{Universal Phone Set ASR}
\label{sec:asr}
We attempt to provide language universal acoustic models by training on many languages sharing a common phonemic representation. We then transfer these models to a new language via a pronunciation lexicon with the same phonemic representation as used in training. We refer to this ASR as universal phone set ASR and we use the same approach as in \cite{wiesner2018jhu}. Following \cite{wiesner2018jhu}, for experiments on Tigrinya and Oromo, we use the same selection of 10 BABEL languages for ASR training ($\sim$600h). For Russian, we use 10h subsets of 21 BABEL languages ($\sim$200h) in training (all except Haitian, Vietnamese, Amharic, Georgian). This reduces training time, provides better phoneme coverage, and performs as well or better in word error rate (WER) as the 10 language ASR on the BABEL Haitian, Amharic and Georgian dev sets.
The final acoustic models are time-delay neural networks (TDNNs, \cite{peddinti2015time}) trained with the lattice-free version of the maximum mutual information (LF-MMI) criterion \cite{povey2016purely}.

During a LORELEI evaluation we also have access to a few hours (2-10) of consultation with a native informant (NI), a native speaker of the IL.
From these interactions we collected an additional 15-30 minutes of IL speech data in both Tigrinya and Oromo. We use this data to adapt the ASR for both languages using the same weight transfer approach as in \cite{manohar2017mgb3}. Since the source languages and ILs use the same phoneset, all layers of the seed neural network (trained on the source languages), including the final layer, were transferred and trained for one epoch on the IL transcribed data.

\vspace{-0.1cm}
\section{Topic Identification}
\label{sec:topicid}
\vspace{-0.1cm}

To leverage the supervised topic annotations of speech segments in multiple dev languages,
we represent each speech segment in all languages as a bag of English words. We derive this representation by building ASR systems to decode the speech and then translate each decoded word into its most likely English translations\footnote{We use the probabilistic bilingual translation tables employed in the MT systems. We use such bilingual lexicons rather than full-blown MT systems to relax the dependency on fully developed IL-to-English MT pipeline that could be unavailable for very-low-resource languages.}.
Support vector machine (SVM) or neural network (NN) based topic classifiers can then be learned by using these English word representations of speech segments in foreign languages along with their associated topic labels.
Thus, using only a translation lexicon, we can always perform topic ID on an IL without its transcribed or topic-labeled speech by using the unadapted universal phone set ASR to decode and translate its speech segments into English words. 
\vspace{-0.05cm}
\subsection{Learning spoken segment representations}
\label{sec:representation}
\vspace{-0.05cm}

Since English word sequences generated using translation tables lack proper syntax, we represent speech segments using a bag-of-words model over the generated English words.
Each speech segment is represented by a vector of unigram occurrence counts over the generated English word sequences and scaled to produce a term frequency-inverse document frequency (tf-idf) feature, which is then normalized to ${\ell}_2$ norm unit length.
Latent Semantic Analysis (LSA) \cite{deerwester1990indexing} transformation can then be learned from the tf-idf features. This transformation effectively merges the dimensions corresponding to words with similar meanings, and maps the high-dimensional tf-idf vectors to a much smaller dimension vector space.

We can also append other auxiliary features to the tf-idf or LSA representations of speech segments. Since our datasets contain segments with music, many of which are out-of-domain, we found that features indicating the substantial presence of music are particularly useful. To generate these features, we build music detectors from the MUSAN dataset \cite{snyder2015musan} and for each speech segment the music detector produces a posterior probability that a substantial portion of music is present.
Denoting the tf-idf/LSA vector as $\textbf{x} \in \mathbb{R}^{d}$, the music posterior as $\delta \in (0, 1)$, and the vector concatenation operation as $\oplus$, our new representation can be created as
$\textbf{x} \oplus \delta$. 

\vspace{-0.1cm}
\subsection{Non-contextual modeling using SVM and NN}
\label{sec:model1}
\vspace{-0.05cm}

Since each speech segment is represented by a vector $\textbf{x}$ and can be associated with one or multiple topics, we perform topic ID by doing multi-label classifications.
The baseline approach is the binary relevance method, which independently trains one binary SVM classifier for each label, and a segment is evaluated by each classifier to determine if the respective label applies.
We use stochastic gradient descent (SGD) based linear SVMs with hinge loss and ${\ell}_2$ norm regularization~\cite{shalev2007pegasos, scikit-learn}. 

Another approach based on feedforward NN\footnote{We simply use NN to refer to multi-layer perceptron in following sections.} is to use an output layer with sigmoid output nodes, one for each label, and train the NN to minimize the binary cross entropy loss defined as
\vspace{-0.2cm}
\begin{equation}
\mathcal{L}(\Theta_{\text{nn}}; \textbf{x}, \textbf{y}) = -  \sum_{k = 1}^{K} ( y_k \log o_k +  (1 - y_k)  \log (1 - o_k)   )
\label{eq:bce}
\vspace{-0.2cm}
\end{equation}
\noindent  where $\Theta_{\text{nn}}$ denotes the NN parameters, \textbf{y} is the target binary vector of topic labels, $o_k$ and $y_k$ are the output and the target for label $k$, and the number of unique labels $K = 12$. 

\vspace{-0.1cm}
\subsection{Contextual modeling using RNN}
\label{sec:model2}
\vspace{-0.05cm}

We explore using recurrent neural network (RNN) to capture the dependencies between context segments. Different RNN variants can be used such as the Elman RNN, long short-term memory (LSTM), or gated recurrent unit (GRU). 
We denote an RNN simply as a mapping 
$\phi: \mathbb{R}^d \times  \mathbb{R}^{d'}\rightarrow  \mathbb{R}^{d'}$ that takes a $d$ dimensional input vector $\textbf{x}$ and a $d'$ dimensional state vector $\textbf{h}$ and outputs a new $d'$ dimensional state vector $\textbf{h}' = \phi (\textbf{x}, \textbf{h}) $.

Consider a spoken document that consists of $n$ spoken segments, as exemplified in Table \ref{tab:sf_eg}. For each $i = 1 \ldots n$, the segment $i$ is represented by a vector $\textbf{x}_i \in \mathbb{R}^{d}$. The document is represented as  $\textbf{X} = \lbrack  \textbf{x}_1 \ldots \textbf{x}_n]$.
We encode $\textbf{X}$ using a bidirectional RNN (BiRNN), and the model parameters $\Theta_{\text{rnn}}$ associated with this BiRNN layer are $\phi_f$,  $\phi_b : \mathbb{R}^d \times  \mathbb{R}^{d'}  \rightarrow   \mathbb{R}^{d'} $.
Thus the segment representation vectors are encoded by forward and backward RNNs as
\vspace{-0.2cm}
\begin{equation}
\begin{split}
 & \textbf{f}_j =  \phi_f (\textbf{x}_j, \textbf{f}_{j-1} )  \quad  \forall j = 1 \ldots n  \\
 & \textbf{b}_j =  \phi_b (\textbf{x}_j, \textbf{b}_{j +1} ) \quad  \forall j = n \ldots 1
\vspace{-0.2cm}
\end{split}
\label{eq:rnn}
\end{equation}
\noindent We assume zero initial state vectors $\textbf{f}_0$ and $\textbf{b}_{n+1}$. And a contextual representation is induced as

\quad $\textbf{h}_i = \textbf{f}_i \oplus \textbf{b}_i    \quad  \forall  i = 1 \ldots n$.

\noindent  We denote the entire operation as a mapping BiRNN$_{\Theta_{\text{rnn}}}$:
 
\quad $(\textbf{h}_1 \ldots \textbf{h}_n) \leftarrow$  BiRNN$_{\Theta_{\text{rnn}}}(\textbf{x}_1 \ldots \textbf{x}_n)$.
 
\noindent Therefore, instead of the non-contextual $\textbf{x}_i$, the contextual $\textbf{h}_i$ is used as input to the feedforward fully connected layers for final classification:

\quad $\textbf{o}_i \leftarrow$  NN$_{\Theta_{\text{nn}}}(\textbf{h}_i)    \quad   \forall i = 1 \ldots n$
 
\noindent where $\textbf{o}_i$ denotes the final output vector. The joint loss 

\quad $ \mathcal{L}(\Theta_{\text{rnn}}, \Theta_{\text{nn}}) = \sum_{i=1}^n  \mathcal{L}(\Theta_{\text{nn}}; \textbf{h}_i, \textbf{y}_i)$ 

\noindent is calculated by Eq. \ref{eq:bce}.

\vspace{-0.1cm}
\subsection{Contextual modeling using attention}
\label{sec:model3}

\begin{figure}[t]
\centering
\includegraphics[width=0.99\linewidth]{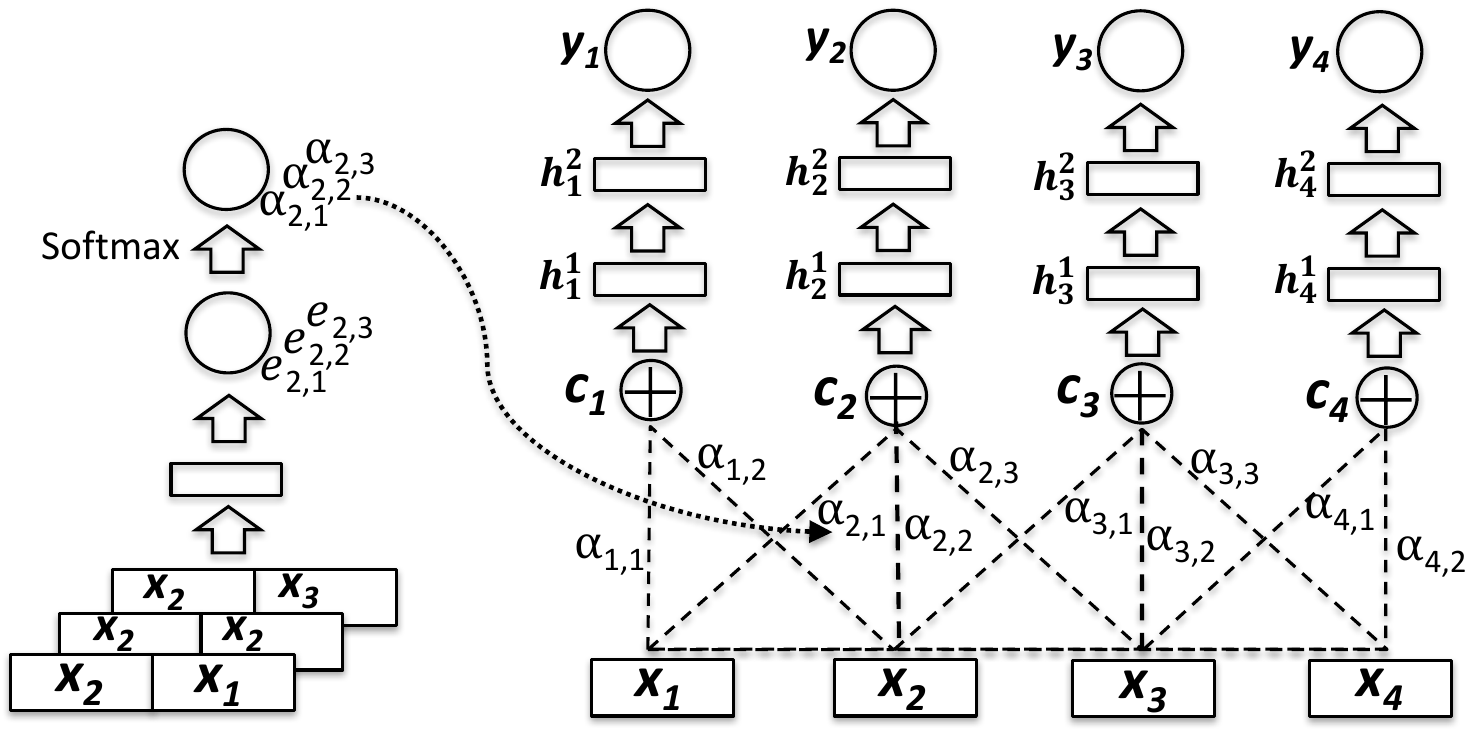}
\vspace{-0.2cm}
\caption{\it Illustration of the proposed contextual modeling using attention, which operates on a spoken document of 4 segments, and leverages each 1-nearest left and right context segments to classify the target $\textbf{x}_{i}$, for each $i = 1 \ldots 4$.}
\label{fig:attn}
\vspace{-0.34cm}
\end{figure}

Consider a spoken document $\textbf{X}$ as above. For each target segment $\textbf{x}_i$,
RNNs implicitly encode its context segments as $\textbf{f}_{i -1}$/$\textbf{b}_{i +1}$, but the RNN non-linear transformations make it hard to control the interaction between segments. 
Instead, we explicitly perform a convex combination of the target and context segments using an attention mechanism \cite{bahdanau2014neural}. 
For each $i = 1 \ldots n$, now consider classifying $\textbf{x}_i$. 
We aim to produce a new contextual vector representation $\textbf{c}_i$ to replace $\textbf{x}_i$, by combining $\textbf{x}_{i}$ and its contexts $\textbf{X}\setminus \textbf{x}_i$. Then each $\textbf{c}_i$ is followed by fully connected layers for final classification as in Section \ref{sec:model1}.
To do so, let $z_i$ be a categorical latent variable with sample space $\{1 \ldots n\}$, which encodes the desired selection among $\textbf{X}$ based on a query $\textbf{q}_i$. 
We let the query be $\textbf{x}_i$ itself, i.e., $\textbf{q}_i = \textbf{x}_i$, since $\textbf{x}_i$ has been produced specifically to encode the semantic information pertaining to segment $i$. 
Then we assume the source position to be selected and attended to follows a distribution, $z_i\sim p(z_i = j | \textbf{q}_i, \textbf{X}),  \forall j = 1 \ldots n$, and therefore the contextual representation $\textbf{c}_i$ is defined as an expectation:
\vspace{-0.3cm}
\begin{equation}
\begin{split}
\textbf{c}_i &  =  \mathbb{E}_{z_i\sim p(z_i | \textbf{X}, \textbf{q}_i)}[ \textbf{x}_{z_i} ]   =   \sum\limits_{j=1}^n p(z_i = j | \textbf{q}_i, \textbf{X}) \ \textbf{x}_j  \\
& =   \sum\limits_{j=1}^n \alpha_{ij} \textbf{x}_j     
\end{split}
\label{eq:attn1}
\end{equation}
\vspace{-0.2cm}
\noindent The weight $\alpha_{ij}$ of each $\textbf{x}_j$ is computed by
\vspace{-0.1cm}
\begin{equation}
\alpha_{ij} =  \frac{ \exp( e_{ij} ) }{  \sum_{k=1}^n \exp( e_{ik} ) },   \quad  \forall  j = 1 \ldots n
\label{eq:attn2}
\end{equation}
\noindent where $e_{ij} = f(\textbf{q}_i, \textbf{x}_j)$, called an alignment model \cite{bahdanau2014neural} that scores how important the segment $j$ is to help classify the query segment $i$. 
We parameterize it with a single-layer NN,
\vspace{-0.2cm}
\begin{equation}
\begin{split}
  e_{ij} & = \textbf{w}^T \sigma (\textbf{W}_{1} \textbf{q}_i + \textbf{W}_2 \textbf{x}_j + \textbf{b}_1) + b_2  \\
           & = \textbf{w}^T \sigma (\textbf{W}_{1} \textbf{x}_i + \textbf{W}_2 \textbf{x}_j + \textbf{b}_1) + b_2, \quad  \forall  j = 1 \ldots n
\end{split}
\label{eq:attn3}
\vspace{-0.4cm}
\end{equation}
\noindent where $\sigma$ is an activation function, and $\textbf{W}_1$, $\textbf{W}_2 \in \mathbb{R}^{d' \times d}, \textbf{w}, \textbf{b}_1 \in \mathbb{R}^{d'}, b_2 \in \mathbb{R}^{1}$ are the weight matrices and jointly learned with all the other NN parameters. 
Note that to classify the target $\textbf{x}_i$, the contexts close to $\textbf{x}_i$ can be more relevant than the distant ones, so we can also use a truncated context window and only consider its $L$/$R$ nearest left/right contexts, i.e., for each $j = \max(0, i- L) \ldots \min(i + R, n)$ in Eq. \ref{eq:attn1}, \ref{eq:attn2} and \ref{eq:attn3}.
The complete modeling framework is illustrated in Figure \ref{fig:attn}, which uses the 1-nearest left and right contexts (i.e. when $L=R=1$).

The intuition behind such process is that, although the overall feature vector $x_i$ may not be salient enough to produce high posteriors for the correct topic labels, certain feature dimensions in $x_i$ are indicative of the correct topics, so that the alignment model of Eq. \ref{eq:attn3} can still capture those informative feature dimensions and give the useful context segments higher scores $e_{ij}$ and higher weights $\alpha_{ij}$. 
The weights are used in a convex combination of Eq. \ref{eq:attn1} such that the useful context features are explicitly combined to produce a contextual representation $\textbf{c}_i$. 

In contrast with the deterministic RNN mapping, the attention mechanism allows for selectively using the contexts in a dynamic manner. 
Consider that, given the left contexts of $\textbf{x}_i$, the forward RNN produces a context vector $\textbf{f}_{i-1}$ as in Eq. \ref{eq:rnn}, and the context vector $\textbf{f}_{i-1}$ is used in a deterministic function $\phi_f (\textbf{x}_i, \textbf{f}_{i - 1})$ regardless of whatever the $\textbf{x}_i$ is. 
However, given different $\textbf{x}_i$, the attention model is able to produce different context weights given different input query vector $\textbf{q}_i$ (since $\textbf{q}_i = \textbf{x}_i$ in Eq. \ref{eq:attn3}); i.e., the contexts will be weighted accordingly for different $\textbf{x}_i$, so that any context can only be effectively used when the attention model detects its relevance and gives it a high weight by Eq. \ref{eq:attn2} and \ref{eq:attn3}. The alignment model (Eq. \ref{eq:attn3}) is explicitly learned as a selector to dynamically detect relevant and useful contexts over irrelevant ones. 

However, as yet, given a fixed input query $\textbf{q}_i$, the alignment model of Eq. \ref{eq:attn3} equally considers the other input features $\textbf{x}_j$, for each $j = 1 \ldots n$, in the attention computation, remaining unaware of that the segment $i$ is being the target one to classify. 
Therefore, inspired by the position-based gating procedure in \cite{chorowski2014end}, the scores $e_{ij}$ can be penalized based on the relative position of the context segment $j$ and target $i$ before being normalized to weight $\alpha_{ij}$:
\begin{equation}
\alpha_{ij} =  \frac{ d(i, j) \exp( e_{ij} ) }{  \sum_{k=1}^n d(i, k)  \exp( e_{ik} ) }, \quad  \forall  j = 1 \ldots n
\label{eq:attn4}
\vspace{-0.08cm}
\end{equation}
where $d(i, j)$ is a gating function of one hidden layer NN and logistic sigmoid output ($[0, 1]$):
\vspace{-0.12cm}
\begin{equation}
    d(i, j) = \left\{
                \begin{array}{ll}
                  1,     \hspace{4.29cm}    j = i    \\
                   \sigma_2 ( w_2 \sigma_1 ( w_1 |  i - j |  + b_1) + b_2),  \quad  \forall  j \neq i      \\
                \end{array}
              \right.
\vspace{-0.12cm}
\end{equation}
\noindent where $\sigma_1$ is an activation function (tanh), $\sigma_2$ a sigmoid function, and $w_1, w_2, b_1, b_2 \in \mathbb{R}^{1}$.
Such additional gating procedure helps favor the weight of target $\textbf{x}_i$ and penalize the effects of any contexts far from the target, so that it can presumably prevent $\textbf{c}_i$ (Eq. \ref{eq:attn1}) from being overwhelmed by context segments regardless of the target $\textbf{x}_i$.

\vspace{-0.12cm}
\section{Experiments}
\label{sec:exp}
\vspace{-0.12cm}
\subsection{Experimental setup}

\vspace{-0.12cm}
\subsubsection{Data}
\label{subsubsection:data}
\vspace{-0.2cm}
\setlength{\tabcolsep}{0.082cm}
\begin{table}[h]
\vspace{-0.3cm}
\centering 
\begin{tabular}{   l  p{2.29cm}     p{0.9cm}    p{0.9cm}      c     p{1.8cm}  }
\hline \hline
               &            Language        &                                               &                                                   &       Topic    &  \quad  ASR         \\
 Dataset &             Pack                &        $|\mathbb{D}_{doc}|$           &       $|\mathbb{D}_{seg}|$     &      Label                &   \ \ Corpora     \\
\hline \hline
\multirow{1}{*}{ }  &   Turkish                    &       212      &    2095       &  LDC  &  BABEL \cite{trmal2014keyword}    \\
 			  &    Arabic                      &       \ \ 47     &     1025        &                  LDC       &    GALE \cite{khurana2016qcri} \\    
                          &      Spanish                   &       198     &     \ \ 393        &             LDC        &  HUB4 \cite{hub4_ne}  \\  
Dev                   &      US English            &       154     &    \ \  842        &       LDC         &    \quad  --    \\  
                         &      Mandarin DEV       &     \ \ 77       &     \ \ 100        &         NI      &  GALE \cite{gale_zh}   \\    
                        &        Tigrinya DEV        &     130     &   \ \ 159            &     NI  &    Universal       \\  
                        &       Oromo DEV           &      241    &   \ \  364           &    NI    &     Universal        \\  
 \hline \hline
                       &      Mandarin EVAL      &   119     &   \ \  724            &   LDC     &    GALE    \cite{gale_zh}             \\  
Eval                &     Russian                  &   136     &  \ \   787              &      LDC     &    Universal     \\ 
                     &      Tigrinya EVAL         &   116     &      1095         &    LDC    &    Universal      \\  
                     &      Oromo EVAL          &    \ \ 46      &   \ \   709          &    LDC     &    Universal     \\  
 \hline \hline
\end{tabular}
\vspace{-0.4cm}
\caption{\label{tab:data} {\it LORELEI speech data description. $|\mathbb{D}_{doc}|$ denotes the number of documents. $|\mathbb{D}_{seg}|$ denotes the number of segments. Manual transcripts are provided for US English corpus.}}
\vspace{-0.4cm}
\end{table}

The dev and eval datasets we used are as shown in Table \ref{tab:data}. 
For Turkish, Arabic, Spanish and English\footnote{Turkish (LDC2016E109), Arabic (LDC2016E123), Spanish (LDC2016E127), and US English (LDC2017E50). Since Spanish set is overwhelmed by the segments of topic ``Elections and Politics", we filtered out all segments that include that topic.}, each language is a single dataset and seen as dev set. Their topic label annotations for all segments are given, and used for training the topic ID classifiers. 

For Mandarin, Tigrinya and Oromo\footnote{Mandarin DEV (LDC2016E108), Mandarin EVAL (LDC2016E115), Tigrinya DEV (LDC2017E35), Tigrinya EVAL (LDC2017E37), Oromo DEV (LDC2017E36), and Oromo EVAL (LDC2017E38)}, each language has one DEV and EVAL set respectively; true topic labels on these DEV sets are unavailable, so we selected some segments, collected their hypothesized topic labels from NI, and included them into the classifier training. 
Also on these DEV sets, we selected some segments for the NI to transcribe and used them for ASR adaptation\footnote{The total given NI session for consultation was 2 hours for Mandarin, 10 hours each for Tigrinya and Oromo. Only on Tigrinya and Oromo DEV sets, we collected transcribed speech from the NI, 27 mins and 18 mins respectively.}. More NI details can be found in \cite{wiesner2018jhu}. 
The EVAL sets of these three languages, in addition to the single Russian dataset\footnote{Russian (LDC2016E111)}, are provided with true topic annotations and are used for evaluating the system performance. 

In sum, when evaluating on Mandarin EVAL or the Russian dataset, the training data for learning topic ID models consists of Turkish, Arabic, Spanish, US English and Mandarin DEV.
When evaluating on Tigrinya EVAL or Oromo EVAL, we use the same training data in addition to Tigrinya DEV or Oromo DEV, respectively.

\vspace{-0.1cm}
\subsubsection{Evaluation metrics}
\vspace{-0.05cm}

Under the LORELEI Speech SF evaluation framework as described in \cite{malandrakis2017extracting}, topic ID system outputs are evaluated in two layers using average precision (AP, equal to the area under the precision-recall curve). 
The \emph{Relevance} layer is to separate the segments with at least 1 in-domain topic from non-relevant out-of-domain segments. 
Specifically, each segment is given 11 posteriors over each in-domain topic, and the Relevance scorer takes the maximum one as the in-domain posterior, and compares it against the true binary label to compute the AP. 
The \emph{Type} layer is to detect all present in-domain topics. Type scorer computes the micro-averaged precision and recall across 11 in-domain topics, and then compute the AP.


\vspace{-0.1cm}
\subsubsection{ASR}
\vspace{-0.05cm}

Audio transcripts exist only for the LORELEI English speech dataset. For the Turkish, Arabic, Spanish and Mandarin datasets, we used preexisting transcribed speech corpora, as shown in Table \ref{tab:data}, to train ASR systems with Kaldi \cite{povey2011kaldi}, and then decoded the LORELEI datasets using the appropriate ASR. 
For Russian, Tigrinya and Oromo, transcribed speech corpora were unavailable and we used the universal phone set ASR to decode each corpus.
For the Tigrinya and Oromo the pronunciation lexicons were obtained as described in \cite{wiesner2018jhu}.
For Russian, we used \texttt{wikt2pron} \cite{wikt2pron} to generate a seed lexicon by scraping Wiktionary for XSAMPA pronunciations of all Russian words found in the provided monolingual text and then proceeded as in \cite{wiesner2018jhu}. We also filtered out all words not written in Cyrillic, and to discard apparent misspellings, we used only the 600k most frequent remaining words.
Note that speech segment lengths vary between 5 seconds and 2 minutes, with an average duration of about one minute.
Since ASR systems have difficulty decoding long segments, we further segmented the audio using either the overlapped segmentation approach as in \cite{peddinti2015reverberation}, or voice-activity-detection (VAD) again as in \cite{wiesner2018jhu}. 
For the overlapped segmentation, we used chunks 15 seconds long repeated every 10 seconds and then filtered the transcripts by removing words whose midpoints were within 2.5 seconds to the chunk edge before combining them into a single transcript.

In addition, we trained two Gaussian mixture models (GMMs) on the speech and music portions of MUSAN \cite{snyder2015musan}. Each speech segment is split into 15 second chunks but without overlap. Then for each chunk, two average frame-level log-likelihoods were calculated by the music and speech GMMs respectively, to further produce a music-to-speech log-likelihood ratio $\gamma$. $\gamma$ went through a sigmoid function and produced a posterior score. Finally for each speech segment, we used the maximum posterior score over all chunks as the music posterior feature $\delta$ for that segment, which was then concatenated to the LSA features (Section \ref{sec:representation}).


\vspace{-0.02cm}
\subsubsection{MT}
\label{sec:mt}
\vspace{-0.02cm}

Supervised topic label information in various languages can all be projected into English topic classifiers through bilingual (i.e., foreign language to English) translation lexicons.  
Each bilingual MT table was derived from the parallel training data with words aligned automatically by the GIZA++~\cite{giza++} and Berkeley aligner~\cite{berkeley_aligner}, independently under the MT effort. 
Any preexisting training data can be used in addition to the data provided by the LORELEI program. 
 
We translated each foreign word in the ASR transcript into its four most likely English translations.  
Then we mapped any unicode data into their nearest ASCII characters, and filtered stop words using the lists from \cite{scikit-learn, bird2009natural}, and any words with three or fewer characters. 
\vspace{-0.1cm}
\subsubsection{Topic ID models}
\vspace{-0.01cm}

First, the tf-idf or LSA features were learned as described in Section \ref{sec:representation}. For the four eval languages overall, we found LSA dimensions over $\{300, 600, 900\}$ can generally produce improvements over tf-idf features, and the ones we finally used are shown in Table \ref{tab:parameter}. 

The non-contextual SVM and NN were learned as in Section \ref{sec:model1}. Contextual RNN and attention based models are described in Section \ref{sec:model2} and \ref{sec:model3} respectively. 
Also, validation data is needed for model parameter tuning and during NN training.
While evaluating Mandarin, we left a small portion out of the training data as validation data. 
While evaluating Tigrinya, Oromo and Russian, we used the Mandarin EVAL dataset as validation data.
The performance of SVMs did not vary much after 30 SGD epochs.
While each NN-based model was trained for up to 50 epochs, the model with the best accuracy on the validation data was used for evaluation on the eval data. 
For each experiment, we repeated it 5 times, and the means are reported in Table \ref{tab:results} (standard deviation is omitted for clarity).

Some parameters were tuned and shared for all languages.
SVMs used  ${\ell}_2$ regularization constant $0.001$ on  tf-idf features. 
All NN-based models had hidden layer size 512 and rectified linear unit (ReLU) nonlinearities, and were trained with Adam optimizer~\cite{kingma2014adam}.
Non-contextual NN used mini-batch size of 256 spoken segments. 
Contextual RNN or attention based models used the mini-batch size of 6 spoken documents. 
For RNN-based models, we found GRU slightly outperformed the conventional Elman RNN or LSTM, and we used the GRU layer that took the LSA features as inputs. 

\setlength{\tabcolsep}{0.04cm}
\begin{table}[t]
\caption{\label{tab:parameter} {\it Differing topic ID model parameters.}}
\centering 
\begin{tabular}{  c  | c |  c   |  c  |    c  }
\hline \hline
 Eval language     &       Russian            &   Mandarin       &      Tigrinya    &      Oromo     \\
\hline 
LSA dimension      &        \multicolumn{2}{c | }{  300 }        &     \multicolumn{2}{c}{  900 }   \\    
 \hline
SVM  ${\ell}_2$  regularization    &   0.001      &         \multicolumn{3}{ | c }{  0.0001 }   \\  
 \hline
\#  hidden layers in NN                &       1         &         \multicolumn{3}{ | c }{ 2 }            \\  
\#  hidden layers in RNN-based   &       0         &        \multicolumn{3}{ | c }{ 1 }            \\ 
\#  hidden layers  in Attn-based    &       1         &        \multicolumn{3}{ | c }{ 2 }            \\    
Dropout rate                                 &       0.5         &     \multicolumn{3}{ | c }{  0.25 }       \\ 
\hline \hline
\end{tabular}
\vspace{-0.5cm}
\end{table}
\setlength{\tabcolsep}{0.129cm}
\begin{table*}[t]
\caption{\label{tab:results} {\it Topic Identification average precision results on LORELEI speech datasets. 
Attention$^1$ or Attention$^2$ are to use 1 or 2 nearest context segments, respectively. 
POS denotes that the additional position-based gating procedure is enabled. 
Last row shows 10-fold cross-validation results on each eval set using ASR transcripts and true topic labels (without using MT or any other dev set), as oracle results for comparison.}}
\vspace{1mm}
\centerline{ 
\begin{tabular}{p{1.99cm}  l |c c|c c|c c| c c | c  c}
\hline \hline
 &           &     \multicolumn{2}{ | c  }{Mandarin }   &     \multicolumn{2}{ |c  }{Russian }   &    \multicolumn{2}{ | c  }{   Tigrinya  }  &     \multicolumn{2}{ |c  }{ Oromo }  &     \multicolumn{2}{ |c  }{ Average }       \\ 
& Model   &               Type   & Rel             &       Type   & Rel                   &    Type   & Rel                         &    Type   & Rel                    &      Type   & Rel       \\
\hline \hline
 \multirow{1}{*}{Non-contextual}   &   tf-idf + SVM                          &    0.458     &  0.702     &   0.382     &   0.854   &  0.371     &    0.554         &  0.382      &     0.772      &   0.398     &  0.721      \\
                             &   \ LSA  + SVM                                                &    0.505     &  0.739     &   0.386    &    0.856   &  0.392     &    0.561         &  0.409     &     0.782      & 0.423    & 0.735   \\ 
                             &  \ LSA  + Music + SVM                                   &    0.510     &  0.742     &    0.408    &   0.870    &  0.422     &    0.600        &  0.423     &    0.822        &  0.441       &    0.759   \\  
                             &  \ LSA + Music + NN     &   \textbf{0.519}  &  \textbf{0.743}  &    \textbf{0.415}  & \textbf{0.881}  &  \textbf{0.451}  &   \textbf{0.625} &  \textbf{0.436} &  \textbf{0.819}  &  \textbf{0.455} &  \textbf{0.767}  \\    
\hline
\multirow{1}{*}{Contextual}  &   \ LSA + Music + RNN                     &   0.525    &   0.737    & 0.430  &   0.894   &   0.389   &     0.578       &  0.467  &   0.820    &   0.453  &  0.757    \\  
                            &  \ LSA + Music + NN + Attention$^1$               &   \textbf{0.544} &  \textbf{0.741}    &   \textbf{0.466}    &   \textbf{0.888}   &  0.407  &   0.597  &   \textbf{0.495}  &  \textbf{0.828}   &  0.478  &  0.764  \\  
                            &  \ LSA + Music + NN + Attention$^1$ + POS    &   0.542    &  0.744  &  0.449  &  0.884  &  \textbf{0.455}    &   \textbf{0.618}       &   0.482   &  0.830    &  \textbf{0.482}  &  \textbf{0.769}  \\  
                            &  \ LSA + Music + NN + Attention$^2$               &   0.537   &  0.742  &   0.461  &   0.892  &  0.365   &   0.557    &    0.494   & 0.838 & 0.464  & 0.757  \\
                            &  \ LSA + Music + NN + Attention$^2$ + POS    &   0.543  &   0.746      &    0.448 &   0.887   &   0.444     &  0.611   &  0.491  &  0.831    &      \textbf{0.482}    &  \textbf{0.769}   \\  
\hline \hline
\textit{Non-contextual}  &  \textit{LSA  + SVM, 10-fold CV}     &   \textit{0.576}  &  \textit{0.843}   &  \textit{0.444}  &  \textit{0.838}  &  \textit{0.574}  &   \textit{0.719}  &  \textit{0.419} &  \textit{0.750}  &   \textit{0.503}  &   \textit{0.788}   \\
\hline \hline
\end{tabular}}
\vspace{-0.6cm}
\end{table*}

The remaining parameters were the same when evaluating Mandarin, Tigrinya and Oromo, but differed for Russian, as shown in Table \ref{tab:parameter}. 
When evaluating Russian, we found using SVM ${\ell}_2$ regularization constant $0.001$ on LSA features, one NN hidden layer and dropout rate $0.5$ gave much better results instead;
presumably because the universal phone set ASR for Russian was unadapted, the resulting transcripts were more noisy and required stronger regularization.
 Also, we used one GRU layer directly followed by the output layer. Each contextual vector $\textbf{c}_i$ (Section \ref{sec:model3}) was followed by one hidden layer instead of two. 
Note that we used the above model parameters different from other three eval languages to obtain optimal results for both Russian non-contextual and contextual models, so that the comparisons between the two are fair. 
In other words, within each eval language, we focus on drawing fair comparisons between its optimal non-contextual and contextual models.

\vspace{-0.08cm}
\subsection{Non-contextual topic ID results}
\vspace{-0.08cm}

Table~\ref{tab:results} first shows the results based on non-contextual model SVM and NN. 
The LSA transformation on tf-idf features substantially improved performance across the board, and also mapped the high-dimensional tf-idf vectors (around 25k) to a dimension small enough for the LSA features to be used as inputs to NN-based models. 
Additionally, appending auxiliary music posteriors (Section \ref{sec:representation}) to the LSA features can produce large gains, except on Mandarin; we found for the Mandarin dataset music was less indicative of out-of-domain topics.
Finally, feedforward NNs were generally more competitive than linear SVMs when using the same input LSA features.

\vspace{-0.07cm}
\subsection{Contextual topic ID results}
\label{subsection:result2}
\vspace{-0.02cm}

Table~\ref{tab:results} further shows the results of our experiments using the proposed contextual RNN and attention models. The GRU-based contextual models outperformed the best non-contextual NN models on Russian and Oromo, but not on Mandarin or Tigrinya. 
For Mandarin, we had a high-performing ASR system trained on about 600 hrs of transcribed speech, so the Mandarin transcripts were much more accurate than other languages, which presumably made it more difficult to improve the non-contextual baseline results; inference from contexts might be helpful to recover the ASR errors in the target segment, and thus better ASR transcripts often allow for confident classification without having to consider additional contexts.
For Tigrinya EVAL set, we found around 72\% of the segments were out-of-domain; i.e., if a target segment is mostly surrounded by out-of-domain segments, using its contexts can give adverse effects, and the overall results can be worse than the context-independent counterparts.

We further experimented with contextual attention based models, using the contexts of 1 or 2 nearest left and right segments, i.e. when $L=R=1$ or $L=R=2$ in Section \ref{sec:model3}. 
The attention-based models outperformed the non-contextual models, except on Tigrinya, due to the overwhelming amount of out-of-domain segments, as discussed above. 
However, we can match the performance of the non-contextual models on Tigrinya, with only a small performance loss in the other languages, by using the additional gating mechanism in Eq. \ref{eq:attn4}. The gating mechanism partially penalizes the context effects and makes the model aware of the target segment location. 
Note that, the attention-based models consistently outperformed the RNN-based models, and it demonstrates the efficacy of the gated attention mechanism that dynamically selects and uses more relevant contexts instead of receiving contexts in a deterministic manner.     

Overall, with respect to the best context-independent models, the contextual attention based models produced comparable performance on Tigrinya, and produced significant performance improvements on the rest three eval languages. Also, the results of using wider contexts, i.e., 2-nearest left and right segments, were comparable to those of using 1-nearest only.
In addition, the attention function we used in Eq. \ref{eq:attn3} is also called additive attention, and we found it outperformed the dot-product (multiplicative) attention \cite{vaswani2017attention}. 
We also experimented with multi-head attention \cite{vaswani2017attention} and component (or multi-dimensional) attention \cite{das2018advancing}, but none of these techniques can give us better results, presumably due to the small size of our topic model training data.

\vspace{-0.05cm}
\subsection{Ten-fold cross validation analysis}
\vspace{-0.05cm}


So far, we have only used English translations of each dev and eval language to resolve the language mismatch, but the training and eval datasets can be severely mismatched.
An oracle result against which we can compare is the 10-fold cross validation (CV) performance on each eval set itself, where each experiment uses part of the true eval set topic labels for training.
For each eval language, we split the corresponding eval set into 10 folds, used the extracted LSA features over the raw ASR transcripts (without translation or any data from other language), completed 10 monolingual supervised SVM classifications with true topic labels, and reported the average of each 10 experiments as shown in the last row of Table \ref{tab:results}. 

For each language, such 10-fold CV results give us estimates of the topline numbers we could achieve with around 700 in-domain training exemplars. First, the gap between each topline number and the full accuracy (i.e. AP = 1)  indicates the given ASR quality and the intrinsic difficulty of each eval dataset. 
Next, comparing our cross-lingual approach with such monolingual topline, we found using the above contextual topic ID approach had reduced the gap on Mandarin, and surpassed the topline on Russian and Oromo, while falling behind on Tigrinya (due to the train-test discrepancy in the amount of out-of-domain segment occurrences as discussed in Section \ref{subsection:result2}).

\vspace{-0.11cm}
\section{Conclusions}
\vspace{-0.08cm}

Audio collected in the wild may contain many topic shifts, and we need to perform topic ID on a sequence of segmented audio.
Each resulting speech segment is of reasonable length and semantically self-contained, such that each of them can be independently classified. 
However, we have performed comprehensive experiments on the LORELEI datasets in a realistic low-resource scenario, and have found that exploiting the context segments can provide significant topic ID performance improvements over the context-independent models.
Finally, comparing our contextual frameworks, we demonstrate that the proposed attention modeling which leverages context segments in a selective approach can consistently outperform the RNN-based alternatives.





\bibliographystyle{IEEEbib}
\bibliography{thesis,refs}

\end{document}